\documentclass[10pt, a4paper]{article}

\usepackage[]{lrec-coling2024} % this is the new style
\usepackage{hyperref}
\usepackage{amsmath}
\usepackage{amssymb}
\usepackage{cleveref}
\usepackage{multirow}
\usepackage{xspace}
\newcommand{\specialcell}[2][c]{%
  \begin{tabular}[#1]{@{}c@{}}#2\end{tabular}}

\newcommand{\ours}{\textsc{PEaCE}\xspace}

\title{\ours: A Chemistry-Oriented Dataset for Optical Character Recognition on Scientific Documents}

\name{Nan Zhang$^*$$^{\clubsuit}$, Connor Heaton$^*$$^{\clubsuit}$, Sean Timothy Okonsky$^\dagger$, \\ {\bf \large Prasenjit Mitra$^{\clubsuit\spadesuit}$, Hilal Ezgi Toraman$^\dagger$$^\ddagger$$^\diamondsuit$}}

\address{$^{\clubsuit}$College of Information Sciences and Technology, The Pennsylvania State University, USA \\
$^{\dagger}$Department of Chemical Engineering, The Pennsylvania State University, USA \\
$^{\spadesuit}$L3S Research Center, Leibniz University Hannover, Germany \\
$^{\ddagger}$Department of Energy and Mineral Engineering, The Pennsylvania State University, USA \\
$^{\diamondsuit}$Institutes of Energy and the Environment, The Pennsylvania State University, USA \\
         \{njz5124, czh5372, sto5087, pmitra, hzt5148\}@psu.edu\\}
\abstract{
Optical Character Recognition (OCR) is an established task with the objective of identifying the text present in an image. While many off-the-shelf OCR models exist, they are often trained for either scientific (\textit{e.g.}, formulae) or generic printed English text. Extracting text from 
chemistry publications
requires an OCR model that is capable in both realms. Nougat,
a recent tool, 
exhibits strong ability to parse academic documents, but is unable to parse tables in PubMed articles, which comprises a significant part of the academic community and is the focus of this work. To mitigate this gap, we present the \textbf{P}rinted \textbf{E}nglish \textbf{a}nd \textbf{C}hemical \textbf{E}quations (\ours) dataset, containing both synthetic and real-world records, and evaluate the efficacy of transformer-based OCR models when trained on this resource. Given that real-world records contain artifacts not present in synthetic records, we propose transformations that mimic such qualities. 
We perform a suite of experiments to explore the impact of patch size, multi-domain training, and our proposed transformations, ultimately finding that models with a small patch size trained on multiple domains using the proposed transformations yield the best performance. 
Our dataset and code is available at \url{https://github.com/ZN1010/PEaCE}. 
\\ \newline  }
\begin{document}

\maketitleabstract
\def\thefootnote{*}\footnotetext{Equal contribution.}\def\thefootnote{\arabic{footnote}}
\section{Introduction}
For documents that are available only as scans, before text processing and natural language processing can be applied, we must convert the images that represent the text in these documents into digital characters before they can be processed to understand their content. In order to do so,
Optical Character Recognition (OCR) is widely
used to extract texts from images in various real-world applications \citep{memon2020handwritten,ye2018unified} and can complement other data 
extraction pipelines \citep{zhang-etal-2022-stapi,wang2020mimic}. Extracting text from images of both scientific texts (\textit{e.g.}, math and physics formulae) and generic printed English plays a vital role in data extraction of scientific articles. A model that is capable in both realms is necessary. Important information in scientific documents is often presented in the form of tables, making data extraction even more difficult.

% For instance, state-of-the-art table extraction and cell recognition models like Multi-Type-TD-TSR \citep{multi-type} work on screenshots of PDF documents and output each table cell as an image. In the scientific domain, the content of each cell can be scientific text or vanilla printed English. So a model that is capable in both realms is necessary.

However, existing open-source OCR models and datasets tend to focus on either scientific texts or generic printed English. Thus, their performance on the documents that contain both 
is suboptimal. For example, Pix2tex \citep{pix2tex},
a pre-trained model that achieves competitive performance on images of math formulae,
is less capable on images containing printed English texts. Tesseract is commonly used to extract vanilla printed English, but it cannot be fine-tuned directly on images of scientific texts, because it outputs plain text strings without formatting for contents such as superscript and subscript~\citep{tesseract}. 
% Furthermore, combining existing datasets from both realms cannot warrant a better-performing model, since scientific notations are not mixed with printed English in any single training image so a model would have a hard time predicting records that contain both type of data. Moreover, labels from each realm suffer inconsistency issue (\textit{e.g.}, OCR datasets of printed English usually do not use markup language like \LaTeX{} to enforce formatting.).

Furthermore, simply combining a vanilla printed English and scientific training corpora is unlikely to yield strong performance on records that are a hybrid of the domains for two reasons. First, there will likely be inconsistencies in the way labels from each corpus are presented - records from the scietific corpus may contain \LaTeX{} formatting not present in the vanilla printed English corpus. While this may be able to be rectified, a second issue still remains - the model will be presented with records containing each type of text separately, but will never be presented with records containing \textit{both} types of text. That is, a model trained on such a corpus will never see scientific text interspersed with vanilla printed English. 

In this work, we seek to address the inability of existing tools
to format text including super/subscripts and other special characters
in academic and scientific papers.
Such text is predominantly printed in English, but often have specially-formatted 
and important characters. For example, 
a document may contain mentions of chemical compounds such as $Na_{2}CO_{3}$.
An OCR model needs to recognize both subscripts, and also 
discern the values in the subscripts as they denote important physical properties of the compound.

Although one may assume it reasonable to segment the text by plain English/special characters and apply domain-specific models on each resulting group, we believe doing so is not the ideal solution. First, doing so would introduce redundant computation into the pipeline. That is, each segment of the image would have to be analyzed twice - once in order to classify the textual content, and again to synthesize it. Furthermore, in doing so, a model would not be able to leverage the temporal dependency in the text. For example, the text ``$Na_{2}CO_{3}\;(Sodium\;carbonate)$'' would get segmented into ``$Na$'', ``$_{2}$'', ``$CO$'', ``$_{3}$'', and ``$(Sodium\;carbonate)$'', and the model would need to synthesize each segment in isolation. We believe a stronger model can be learned processing the entire record at once, leveraging the temporal dependency between the chemical compound, $Na_{2}CO_{3}$, and its name in English, $Sodium\;carbonate$.

Nougat, a newly released model, can perform OCR on entire pages of academic documents, including parsing tables,
% present therein, 
but struggles \textit{significantly} in parsing tables from documents published in PubMed\footnote{\url{https://www.ncbi.nlm.nih.gov/pmc/}} \cite{blecher2023nougat}. As the authors describe, PubMed papers often present tables as embedded images.
So they could not access the ground-truth text present therein without an expensive annotation process. 
% {\color{red} So what? So, they did not train on Pubmed and thus it does not work very well with PubMed documents? If so state that. Here the logical conclusion is sort of unsaid.}
As such, the model often fails to recognize tables in such documents, and when tables are recognized, they are rarely parsed correctly. 
% PubMed hosts papers for a large portion of the life sciences and biomedical community, leaving a large portion of the academic papers unaddressed or under-addressed. 
PubMed hosts a large portion of papers in the life-sciences and biomedical domains, leaving a significant portion of the academic community unaddressed or under-addressed. 

We aim to address the above shortcomings by proposing a new data resource, since there does not exist an OCR dataset that contains images of both scientific texts and printed English to the best of our knowledge. Thus, we introduce \ours (\textbf{P}rinted \textbf{E}nglish \textbf{a}nd \textbf{C}hemical \textbf{E}quations) dataset, containing synthetic and real-world images of text from academic articles, with a particular focus on chemistry papers. Each record in \ours is intended to resemble a cell in a table that may appear in an academic document (a handful of words across two or three lines), and the code we release exposes parameters that make it easy for researchers to generate records of any length and format they desire. A model trained on \ours could then be combined with a state-of-the-art (SOTA) table parsing model such as Multi-Type-TD-TSR \citep{multi-type} to parse the content of tables identified in scientific documents, addressing the market Nougat cannot. 

\begin{figure}[!ht]
\begin{center}
    \includegraphics[scale=0.4]{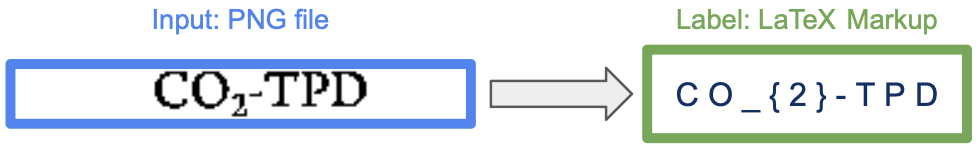}
    \caption{
    A data instance of \ours. 
    % The blue box displays the original screenshot, and the green box contains \LaTeX{} markup as the label of the screenshot.
    The source image is outlined in blue, the \LaTeX{} label in green.
    }
    \label{example}
\end{center}
\end{figure}

\ours has two parts: 1) synthetic records and 2) real-world records. The synthetic portion of the dataset contains 1M images of printed English text, 100k images of numerical artifacts, and 100k images of (pseudo-)chemical equations, subset in mutually exclusive training/dev/test splits. The real-world test set comprises 319 carefully curated records and assesses the performance of OCR models on text from actual chemistry scholarly papers. Figure~\ref{example} shows a data instance from the real-world test set of \ours. All labels in \ours are \LaTeX{}\footnote{\url{https://www.latex-project.org/}} markup as it is a versatile typesetting tool that can express the multitude of non-alphanumeric characters often found in academic papers
such as ``$\sum$'', ``$+$'', and ``$\mathbb{R}$'', and can format super/subscripts. Given that the real-world records contain corruption not found in synthetic records, we propose three transformations - pixelation, bolding, and white-space padding - to mimic these artifacts. % during training. 

% Besides constructing \ours, we also conduct multiple surveys about how state-of-the-art OCR models perform when they are trained and tested on this newly created dataset. Our findings can be broadly summarized into three folds. First, we find that selected models suffer from a significant performance drop on the real-world test set even when they reach competitive scores on test sets from other related domains. This reinforces our motivation of proposing \ours as a dataset for real-world model evaluation on multi-domain testing, because there is no ``hybrid'' dataset from scientific articles that reveals this model weakness to the best of our knowledge. Although \ours is chemistry-oriented, it may be used for other subjects (e.g., math and physics scholarly papers) that involve both domains as well. Additionally, for transformer-based models, we find that the patch size parameter can significantly impact model performance and a small patch size generally leads to better performance. 

% Finally, we see that models trained on multiple-domains do not necessarily outperform models trained on a single domain when applied to said single domain. However, the performance of the multi-domain models greatly surpasses that of single-domain models when evaluated on both domains, highlighting the utility of \ours for training multi-domain OCR models.

We explore how two different versions of the Vision Transformer (ViT) perform when trained on \ours, and our findings can be summarized in three folds. First, although all models perform well on the synthetic tests sets, they exhibit a sharp decline in performance on the real-world test set. This reinforces our motivation for proposing \ours, as existing datasets to not reveal this shortcoming. Then, we observe that the patch size parameter has a significant impact on resulting performance, with smaller patches leading to better performance. Further, we see that models trained in a multi-domain setting, \textit{i.e.} both \ours and im2latex-100k, yield better performance in each domain than a model trained on a single domain. Finally, we also observe that our proposed image transformations improves performance in two of our three test datasets.

This paper mainly contributes the following:
\begin{enumerate}
  \item We propose a novel dataset that contains images of both scientific texts and printed English for training and testing OCR models on articles from the hard sciences, with an emphasis on Chemistry. 
  \item We demonstrate how models that perform well on related datasets perform significantly worse on our \ours real-world dataset, highlighting the value of this new resource.
  \item We present a set of quantitative evaluations to show the effect of patch size, multi-domain training, and image transformations.
\end{enumerate}

\section{Related Work}
% We discuss previous work related to ours such as datasets for training OCR models, deep learning models for computer vision in general, and deep learning models specifically designed for OCR.

\subsection{``Hybrid'' Dataset for OCR}
We did not find any relevant datasets that contain images of both scientific texts and printed English for OCR models on scientific documents. Therefore, we list the recent efforts here that are closest to this paper. \citet{zharikov2020ddi} proposed DDI-100, a dataset of distorted document images. Since the labels of DDI-100 are text strings with corresponding locations, 
this work concentrates vanilla printed English (\textit{i.e.} no formatting for super/subscripts) and 
is not as effective on scientific documents. 
% Moreover, the images in our dataset have much less textual content than DDI-100. 
Furthermore, this dataset contains images of entire documents, but we are primarily concerned with recovering text from smaller scope images without figures (\textit{e.g.}, individual table cells).

\citet{8978078} proposed a large table recognition dataset,
which they dubbed TABLE2LATEX-450K from scientific documents.
Its images contain complete tables whereas ours is focused on cells in tables in images.
It is infeasible to match cell contents (in \LaTeX{} format) with their corresponding pixels in the table images.
Thus, we cannot create a dataset of images that contain scientific text. In addition, work has been done to collect photographs of random academic papers under factors 
such as non-uniform lighting, strong noise, sharpening, skew, and blur \citep{8978020}.
% In this paper, these factors do not matter, because all images are screenshots collected from academic sources.
For our purposes, artifacts such as sub-optimal lighting are not overly important as the models will be presented with images from PDF, or sometimes \textit{scanned} documents, which may incur their own distinct set of artifacts. 

\subsection{Vision Transformer}\label{sec:vit}
\begin{figure}[htb]
    \centering
    \includegraphics[scale=0.375]{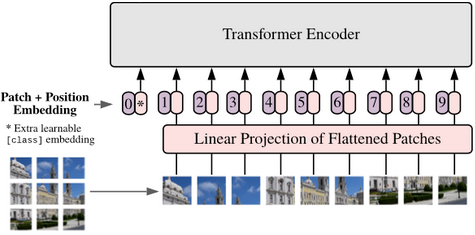}
    \caption{
    Patch projection in the ViT.
    }
    \label{fig:vit_patch_proj}
\end{figure}

Transformers have been used in
a variety of disciplines, including in computer vision (CV), where the variant was simply dubbed the Vision Transformer (ViT) \citep{dosovitskiy2020image}. The ViT retains many of the features of the original transformer designed for machine translation but processes a sequence of image patches instead of token embeddings. As described in Figure \ref{fig:vit_patch_proj}, images are first segmented in to $P*P$ patches of non-overlapping pixels before each patch is projected $\mathbb{R}^{P*P} \Rightarrow \mathbb{R}^{D_M}$ where $D_M$ is the internal dimension of the model.

% As described in Figure \ref{fig:vit_patch_proj}, images are first segmented in to $P*P$ patches of non-overlapping pixels. In the original paper, the authors explore a \textit{base} model where $P=16$ pixels and a \textit{huge} model where $P=14$ pixels. Then, each patch is \textit{flattened} to be a one-dimensional vector and projected to the internal dimension of the model. That is, the dimension of each patch embedding is projected $\mathbb{R}^{P*P} \Rightarrow \mathbb{R}^{D_M}$ where $D_M$ is the internal dimension of the model.

% [Attention]
% When comparing models with patch sizes of $32*32$, $16*16$ and $14*14$, the authors note that ``... the transformer’s sequence length is inversely proportional to the square of the patch size, thus models with smaller patch size are computationally more expensive'' but also that decreasing the patch size and thus increasing the effective sequence length shows surprisingly robust performance improvements. In tandem, the comments suggest that smaller patch sizes allow for better performing models at the cost of computation. 
After exploring patch sizes of $32*32$, $16*16$ and $14*14$, the authors of ViT ultimately conclude that smaller patch sizes allow for better performing models at the cost of computation. \citet{strudel2021segmenter}, reinforce this conclusion, observing ``... the performance is better for large models and small patch sizes''. 
They primarily investigated patch sizes of $32*32$ and $16*16$, only training a single model with the smaller $8*8$ patch size. For the $8*8$ model that was trained, they found conflicting results: one performance metric shows improvement while the other a degradation. 

\citet{than2021preliminary} explored patch sizes ranging from $16*16$ to $256*256$ when classifying chest x-rays by whether or not the patient had COVID-19, with best performance achieved using a patch size of $32*32$. In general, the community has seemingly adopted a patch size of 16*16 as the \textit{de facto} standard for the ViT, in line with the originally proposed \textit{ViT-Base}.

% \citet{than2021preliminary} explored patch size when classifying an individual as having or not having COVID-19 by analyzing their chest x-rays. After exploring patch sizes ranging from $16*16$ to $256*256$, the best performance was achieved using a patch size of $32*32$. In general, the community has seemingly adopted a patch size of 16*16 as the \textit{de facto} standard for many tasks/models in CV, in line with the originally proposed \textit{ViT-Base}.

% [Attention]

\subsection{Pix2Tex}\label{sec:pix2tex}
Pix2Tex \citep{pix2tex} is a competitive OCR model for scientific text that employs both a ViT encoder and decoder. Its ViT encoder comes with a ResNet backbone \citep{resnet}, which means that several ResNet layers are adopted to extract features from source images that are then fed into the ViT encoder. That is, the ResNet backbone is used in place of a patch-projection module. 
% The training data of this model is a combination of im2latex-100k and other sources of math formulae collected from the web, so it was originally trained to perform OCR on images of scientific text. 
% Their purpose was performing OCR on scientific text. 
The training data for the released model checkpoint was a combination of im2latex-100k and math formulae collected from other sourcs on the web.
By training Pix2Tex from stracth on \ours, we are able to evaluate the capability of this architecture on scientific text, printed English, or both.

\subsection{Tesseract}
Tesseract \citep{tesseract} is a popular OCR model that was originally trained for generic printed English. By fine-tuning it on \ours under a multi-domain training setup, Tesseract can possibly detect specicial characters such as super/subscripts on scientific documents by incorporating more vocabulary. The core of Tesseract is an LSTM network \citep{hochreiter1997long}. However, since there was a significant performance gap between Tesseract and Pix2Tex in our preliminary experiment, we do not pursue Tesseract as a baseline in this paper.

% \subsection{MI2LaTeX}
\subsection{Math OCR}
MI2LS is currently the best performing model we are aware of on the im2latex dataset \citep{wang2021translating}. 
% , achieving a BLEU score of 90.28, edit distance of 92.28, and exact match (EM) score of 82.33 \cite{wang2021translating}. 
The model consists of a CNN encoder and RNN decoder, augmented with attention \citep{bahdanau2014neural}. Training the model consisted of two phases: token level and sequence level. 
During token level training, training is based on traditional maximum likelihood estimation (MLE). That is, the model predicts the token most likely to be present at each timestep. During sequence level training, the model is further trained using a reinforcement learning (RL) scheme designed to optimize reward for the emitted sequence as a whole. Here, the reward is an increase in the BLEU score \citep{papineni2002bleu}. This results in a slightly different training objective than during token-level training. The authors ultimately
found that performing sequence level reinforcement learning after token level training did improve the quality of the model, but not to a tremendous extent.

Other efforts towards OCR for math often employ a similar architecture, such as the CNN-LSTM architecture proposed by \citet{mirkazemy2022mathematical} or the U-net architecture proposed by \citet{ohyama2019detecting}. Very recent efforts have began exploring how the transformer architecture can be applied to the OCR task, such as \citet{zhao2021handwritten} exploring how the transformer can be used to synthesize text in handwritten mathematical expressions. MathPix is purported to be proficient at this task, but is a commercial offering with unclear implementation as is thus not considered in this study \cite{Mathpix}.

% They first process the given image via a convolutional feature-extractor, passing the resulting image to a transformer-based encoder-decoder, with the decoder consisting of bi-directional, autoregressive transformer layers. They ultimately find that the transformer-based approach improves performance by 5 to 10 percentage points over various CNN-/RNN-based approaches.

\subsection{Nougat}
Nougat is a recently proposed model for neural optical understanding for academic documents \cite{blecher2023nougat}. Trained on 8.2M pages of academic documents, primarily from arXiv, the model takes as input a PDF page and outputs the identified text, including tables, in a markup language. While the training corpus does contain documents from PubMed Central\footnote{\label{pmc_footnote}\url{https://www.ncbi.nlm.nih.gov/pmc/}} (PMC), such documents typically present tables as images, so Nougat's pre-processing pipeline is unable to identify the text therein when creating a ground-truth record. Accordingly, although the model performs well in the general academic domain, it struggles when parsing tables from PMC, and often fails to recognize them altogether. Seeing as PMC hosts articles for a significant portion of the scientific community, an approach to filling this gap is of value. The introduction of \ours provides a high-quality training resource for this type of document.

\section{\ours Dataset}
% A little bit about how the data we needed didn't exist previously?
To mitigate the 
% long-lasting 
bottleneck of not having a good OCR dataset that contains images (and corresponding labels) of both scientific texts and printed English, \ours contains 1M images of printed English text, 100k images of numerical artifacts, and 100k images of pseudo-chemical equations, along with their \LaTeX{} labels. Furthermore, for an understanding of real-world performance, the dataset also contains 319 images from real-world scientific documents, again with \LaTeX{} ground-truth strings, to serve as a real-world test set. A summary of the records in \ours is presented in Table \ref{tab:dataset_summary}. Scripts to generate the dataset are included in our code release so practitioners can generate different versions of the dataset as they see fit, modifying both the corpus that is sampled to generate the records and the formatting applied to the sampled text.

\begin{table}[htb]
    \centering
    \begin{tabular}{|c|c|}
        \hline
        \multicolumn{2}{|c|}{\textbf{Printed English}} \\
        \hline
        \# Total Characters & 33M \\
        \hline
        \# Unique Characters & 405 \\
        \hline 
        Avg \# Characters / Record & 32.92 \\
        \hline
        \# Records & 1M \\
        \hline
        
        \hline
        \multicolumn{2}{|c|}{\textbf{(Pseudo) Chemical Equations}} \\
        \hline
        \# Total Characters & 7.9M \\
        \hline
        \# Unique Characters & 101 \\
        \hline 
        Avg \# Characters / Record & 78.72 \\
        \hline
        \# Records & 100k \\
        \hline

        \hline
        \multicolumn{2}{|c|}{\textbf{Numeric Records}} \\
        \hline
        \# Total Characters & 1.9M \\
        \hline
        \# Unique Characters & 51 \\
        \hline 
        Avg \# Characters / Record & 18.83 \\
        \hline
        \# Records & 100k \\
        \hline
        
        \hline
        \multicolumn{2}{|c|}{\textbf{Real-world Test}} \\
        \hline
        \# Total Characters & 5,286 \\
        \hline
        \# Unique Characters & 101 \\
        \hline 
        Avg \# Characters / Record & 16.57 \\
        \hline
        \# Records & 319 \\
        \hline
        
        \hline
    \end{tabular}
    \caption{Summary statistics of \ours.}
    \label{tab:dataset_summary}
\end{table}

As stated above, the labels accompanying each record are in the form of \LaTeX{} strings. To be rendered in a \LaTeX{} environment, however, the strings need to be surrounded by ``\$'' characters - \textit{i.e.} in \textit{math mode}. In all cases, the python library \texttt{matplotlib} \citep{hunter2007matplotlib} was used to render text as it has the ability to render \LaTeX{} strings, the format in which all of our ground-truth labels are presented.

\ours exposes an OCR model to a substantial amount of special characters related to chemistry domain. Artifacts such as benzene rings and cubane cube are not included, since they contain many visual artifacts typically not presented in-line with other text and recognizing them does not fall under the umbrella of OCR. While existing datasets may contain similar components of our dataset (\textit{e.g.}, printed english, latex characters, and chemical equations), \ours is the first to integrate these components on a record level. 

\subsection{Printed English Records}
To construct our synthetic printed English records, we repurpose the arXiv\footnote{\url{https://arxiv.org/}} and PubMed\textsuperscript{\ref{pmc_footnote}} datasets originally proposed for long-document abstractive summarization \cite{cohan2018discourse}, and supplement them with a crawl of chemRxiv\footnote{\url{https://chemrxiv.org}} abstracts via the \texttt{paperscraper}\footnote{\url{https://github.com/PhosphorylatedRabbits/paperscraper}} python package. In total, the aggregate dataset used to create our synthetic records contains 100M words from 31k papers. This dataset is then used to create (rendered text, \LaTeX{} ground-truth) pairs.

% To construct our set of printed English training records, we re-purpose the long-document abstractive summarization dataset originally introduced by \citet{cohan2018discourse}. The dataset contains the \LaTeX{} source for over 215k academic papers appearing on arXiv, with an average length of almost 5,000 words. Although not the original intention, this dataset contains an ample amount of text from which we can create (rendered text, \LaTeX{} ground-truth) pairs. % (each training pair consists of a rendered text image and a \LaTeX{} ground truth).  

Specifically, we randomly sample an academic document from which to sample text, and then select a sample of up to $w=10$ consecutive words to serve as the ground truth text. It is uncommon to find \textit{exclusively} vanilla printed English in scientific documents, so we perturb the ground truth text to make it more realistic. First, we randomly add a superscript/subscript to a word in the sampled text with probability of $p_1=3.75\%$ and $p_2=1.25\%$, respectively. Next, \LaTeX{} characters ($\phi$, $\infty$, $\sum$, $\prod$, \textit{etc.}) are randomly inserted to the sequence with probability of $p_3=15\%$. For \LaTeX{} characters that require \textit{arguments} (\texttt{\textbackslash bar\{\}}, \texttt{\textbackslash dot\{\}}, \textit{etc.}), English characters are sampled 50\% of the time, integers the rest. Finally, up to four carriage returns (line breaks) are inserted into sequence with probability $p_4=15\%$. The weight associated with inserting $i$ carriage returns is computed as $\frac{1}{i^2}$.

% Firstly, we randomly add a superscript to a word in the text with probability of 2.5\%, indicative of a footnote in a real-life document, for example. Subsequently, we randomly insert up to four carriage return (\textit{i.e.} new line) tokens in the text. 
% Finally, once the text is sampled and perturbed, one of four fonts is randomly selected and used to render the text to ensure variation among printed characters. 
% Finally, once the text to be included in the dataset has been obtained, one of four possible fonts is randomly selected and used to render the text. This is done to increase the variety within the dataset, exposing the model to the same set of characters in different styles.
% The python library \texttt{matplotlib} \cite{hunter2007matplotlib} is used to render the text. 

Once the text is sampled, it is rendered using one of eight randomly selected fonts in one of six randomly selected font sizes. This process is completed $n=1M$ times to assemble the printed English section of \ours. The python library \texttt{matplotlib} \cite{hunter2007matplotlib} is used to render the text. We would like to note that all of the parameters mentioned above - $w$, $p_1$, $p_2$, $p_3$, $p_4$, and $n$ - are exposed to the end user such that they can easily modify the record generation process. Although not exposed as command line parameters, the potential fonts and font sizes can also be easily modified. 

\subsection{(Pseudo-) Chemical Equation Records}
We also create 100k pseudo-chemical equations such that our model will be exposed to this \textit{form} of language during training. These constructed chemical equations are likely to not abide by the laws of chemistry. However, in order to perform OCR on scientific documents, they are useful for exposing our model to this \textit{structure} of text (\textit{i.e.} sequences of 1-2 characters with subscripts interspersed). We do not pursue existing chemical databases to create chemical compounds and equations, because there does not exist a database that renders different chemical compounds/equations in markup languages (\textit{e.g.}, \LaTeX{}) to the best of our knowledge.

\begin{figure}[htb]
    \centering
    \includegraphics[scale=0.2]{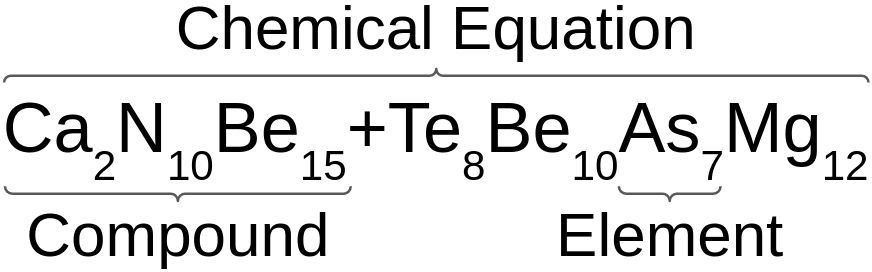}
    \caption{
    Example pseudo-chemical equation.
    }
    \label{fig:chem_eq_ex}
\end{figure}

% We identified 20 different elements that will be used in creating chemical equations in this study. 
To create a pseudo-chemical equation, we first randomly sample a number $n_{compound}$ from 1 to 4 to determine the number of compounds that will be used in the equation. Then, for each compound, we randomly sample a number $n_{elements}$ from 1 to 4 to determine the number of chemical elements in said compound. As elements are randomly sampled to construct the compound, a value $n_{quantity}$ ranging from 1 to 500 is randomly sampled as that elements quantity in the compound. Finally, conjoiners ``+'', ``with'', ``and'', and ``plus'' were randomly sampled to join the constructed compounds. Again, we randomly sample one of eight fonts and one of six font sizes to render the image to increase the variety of characters the model sees during training. An example of the resulting chemical equation record is presented in Figure \ref{fig:chem_eq_ex}. Similar to printed English records, the key parameters ($n_{compund}$, $n_{elements}$, and $n_{quantity}$) are exposed to the programmer.

\subsection{Numeric Records}
\begin{figure}[htb]
    \centering
    \includegraphics[scale=0.3]{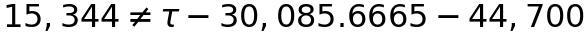}
    \caption{
    Example numerical record.
    }
    \label{fig:numeric_record_ex}
\end{figure}

Numeric records are constructed following a similar process. First, we randomly sample a number $n_{numerals}$ from 1 to 4 to determine the number of numerals that will be created. For each numeral being created, decimal in the range of $0$ to $100,000$ is sampled with probability $p_1=0.5$. Otherwise, a \LaTeX{} math symbol (``$\lambda$'', ``$\beta$'', ``$\Psi$'', \textit{etc.}) is randomly chosen. Each numeral is then \textit{joined} using characters such as ``$+$'', ``$\pm$'', and ``$\neq$''. These records may not abide by all mathematical laws and properties, but they serve to expose the model to this type of notation. This process is performed 100k times, and records are rendered using one of eight fonts in one of six font sizes. Again, $n_{numerals}$ and $p_1$ are exposed to the user as parameters, while the conjoining symbols and set of sampled \LaTeX{} math symbols can be easily modified within the code. An example record is presented in Figure \ref{fig:numeric_record_ex}.

\subsection{Real-World Test Set} \label{3.3}
% \begin{enumerate}
%     \item Obtain chemical scholarly papers
%     \item Parse tables using 
%     \item Identify 250 records as representative of set
% \end{enumerate}
% Nuances in the data - content isn't always ``top left''

To obtain records for our real-world test set, we first collect published scholarly papers relating to polymer pyrolysis and identify 21 tables therein. Then, we transform the PDF pages that contain these 21 tables to images using python library \texttt{pdf2image}. Finally, we pass the images through the Multi-Type-TD-TSR model \citep{multi-type} which will section the original image \textit{by table cell}. That is, the model will emit images containing the content of each cell in the table.

Once we have images describing single cells, we identify a group of 319 cell images that are representative of the overall set. This group contains two subgroups: normal cells and special cells. Normal cells contain text that is written using only alphanumeric characters while special cells contain text that cannot be written in \textit{only} alphanumeric characters and require special \LaTeX{} symbols to be rendered.

Note that this real-world test set contains nuances in the data not found in the other synthetic parts of the dataset. To start, some of the original polymer pyrolysis papers appear to be scanned and are not \textit{true} PDF documents. Thus, some cells may have artifacts such as pixelation or smearing. Furthermore, our synthetic records crop the images such that the text \textit{always} appears in the top-left portion of the image. The same cannot be said for images processed and cropped by Multi-Type-TD-TSR, however. We find that in many cases, the text begins in the \textit{middle} of the image, not the top-left. In addition, the Multi-Type-TD-TSR model sometimes crops incorrectly, resulting in a significant amount of white space between text. These nuances make it a valuable representation of real-world data, because the tools that crops images are not expected to be error-free.

Finally, we note that 33 images in the real-world test set are over 700 pixels wide and are too big to be processed by the models trained in this study. We include these images in the released version of \ours so they would be available to other researchers if desired, but they are not reflected in the performance metrics on the real-world test set presented below.

\section{Experiments}
In this section we describe the experiments that we perform in this study. We first describe our experiments on patch size, identifying nuances of this application perhaps not found in other applications of the ViT, and then our experiments related to multi-domain training for domain-specific applications. We also experiment with different record transformation techniques. We explore the performance of a vision transformer for OCR, dubbed OCR-ViT, as well as Pix2Tex, an OCR-ViT model with a ResNet encoder in place of patch-projection (Section~\ref{sec:pix2tex}). Specifically, the OCR-ViT model consists of a bidirectional encoder to process the source image and an autoregressive decoder to synthesize the text.

\subsection{Effect of Patch Size}\label{sec:impact_of_patch_size}
\begin{figure}[ht!]
    \centering
    \includegraphics[scale=0.295]{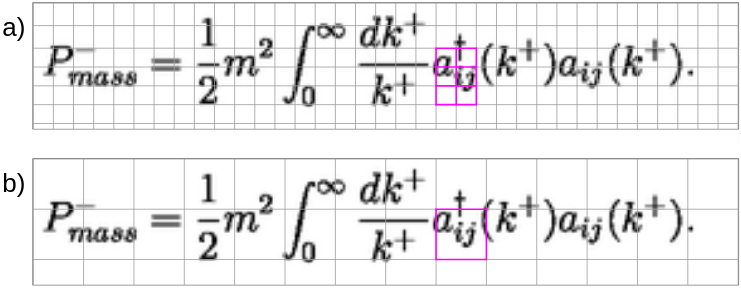}
    \caption{
    a) Example of how a patch size of 8 manifests on an example image. b) Example of patch size 20 on the same image.
    }
    \label{fig:patch_size_comp}
\end{figure}

As mentioned above, we are interested in exploring the impact of the patch size parameter on the resulting performance of a transformer-based OCR model. To this end, we explore OCR-ViT models trained with different patch sizes: $10*10$ and $16*16$. Given images of dimensions
$160*600$, this translates to an effective sequence length of 900 and 375, respectively. To visualize the impact of the patch size in our application, consider Figure \ref{fig:patch_size_comp}, where we compare how different patch sizes manifest on the same source image.

In Figure \ref{fig:patch_size_comp}, we see that there is not only a tradeoff between patch size and effective sequence length, but also between the patch size and the \textit{amount of text} contained in each patch. For example, consider the patches highlighted in pink, which are the ones required to (mostly) describe the portion of the image containing the text ``$a_{ij}$''. When a patch size of 20 is used, all three characters must be described using a single patch embedding. With a patch size of 8, however, six different patches are used to convey the same information.

Recall from \cref{sec:vit} that each patch is presented to the model as a $D_M$-dimensional embedding, where $D_M$ is the internal dimension of the model. In other words, the content of each patch must be described using $D_M$ numbers. When using a $20*20$ patch, then, the text $a_{ij}$ must be described using $D_M$ numbers. For a $8*8$ patch, however, $6 \times D_M$ numbers can be used to describe the same text \-- in some sense, lowering the burden placed on the model. Of course, this comes with a tradeoff of higher computational overhead. 

We hypothesize that the impact of patch size is more pronounced in OCR applications than other applications of the ViT. In our application, the precise content of each patch is of utmost importance - the model must be able to differentiate between an ``e'' and ``c'' or ``i'' and ``j'' for example. In other ViT applications, however, the precise contents of each patch may not be as important - describing the predominant shape and/or color of the content therein, for example. For this reason, we believe exploring the efficacy of a smaller patch size is 
important.

\subsection{Multi-Domain Training}
% In addition to training our models on our newly constructed PEaCE dataset, we also incorporate the im2latex-100k dataset - a dataset containing 100k+ \LaTeX{} equations and their rendered equivalent.
In addition to exploring patch size, we also explore the impact of adding multi-domain training data for domain-specific inference when sufficient domain-specific training data is available. That is, we explore whether training our models on the joint of im2latex-100k \textit{and} \ours yields better performance on the 
test set of im2latex-100k (or \ours) than training on im2latex-100k alone. This kind of experiment can reveal the utility of \ours.

\subsection{Record Transformations}
The synthetic portion of \ours contains \textbf{only} high-resolution records, while the real-world test set contains records with various artifacts - pixelation, smudging, \textit{etc}. We also observe that some records taken from particularly old papers tend to have a very dark, thick font which models in some of our preliminary experiments had difficulty processing. Furthermore, we notice that while the Multi-Type-TD-TSR model does a good job of extracting table cells, it often leaves a non-negligible amount of white space around the textual content of each cell. 

\begin{figure}[ht!]
    \centering
    \includegraphics[scale=0.74]{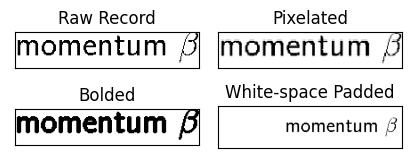}
    \caption{
    Example transformations.
    }
    \label{fig:transforms_ex}
\end{figure}

To address this, we propose three transformations for application on training records: 1) \textit{pixelation}, 2) \textit{bolding}, and 3) \textit{random white-space padding}. An example of these transforms are presented in Figure \ref{fig:transforms_ex}. The \textit{pixelation} filter first compresses the model to a smaller dimension before \textit{expanding} it back to the original dimension, interpolation missing pixels which introduces a small amount of noise. At a high level, the \textit{bolding} filter operates on a pixel-wise basis on a binary, black and white image. If a white pixel has $\geq n$ ``hot'' pixels in a \textit{n-pixel} radius (horizontally, vertically, diagonally), the white pixel is converted to a black pixel. The \textit{random white-space padding} is very straightforward, adding a random amount of white-space along either side of the image. 
% Pseudocode for each transformation is presented in the appendix. 

\section{Results}
% In this section, we present the results from performing the experiments described above.

\subsection{Metrics}
The reported performance metrics are BLEU-4 score, edit distance, and exact match percentage. BLEU score, a precision-based metric, evaluates the similarity between two pieces of text by counting the number of N-grams in a generated sequence that are present in the ground-truth \citep{papineni2002bleu}. BLEU-4 analyzes the precision of uni-, bi-, tri-, and quad-grams in the generated sequence with respect to the ground-truth, giving equal weighting to each. Values range from 0 to 100 where higher is better.

Levenshtein distance is the number of single-character edits that must be made to the strings in order for them to match~\citep{levenshtein1966binary}. 
Following common practice in the domain \citep{wang2020pdf2latex,wang2021translating}, we present Edit Distance (Edit) between a set of $N$ reference strings $R$ and parallel set of $N$ hypothesis strings $H$ as described in Equation~\ref{eq:edit_distance}. Values range from 0 to 100 with higher values indicating a stronger model. 

\begin{equation}
    Edit(R,H) = 1 - \frac{TotalLevenshtein(R,H)}{TotalLength(R,H)} 
    \label{eq:edit_distance}
\end{equation}
where
\begin{align*}
    TotalLevensht&ein(R,H) =\\ &\sum_{i=1}^{N}LevenshteinDistance(r_i,h_i)
\end{align*}
\vspace{-20pt}
\begin{align*}
    TotalLength&(R,H) = \sum_{i=1}^{N}max(len(r_i), len(h_i))
\end{align*}

The exact match (EM) describes the percent of generated hypotheses that match their corresponding ground-truth string \textit{exactly}. EM essentially describes the record-level accuracy. As with the other metrics, values range from 0 to 100 with higher values indicating a stronger model. 

\setlength\tabcolsep{3.6pt}
\begin{table*}[ht!]
    \centering
    \begin{tabular}{|c|c|c|c|c|c|c|c|c|c|c|}
        % \hline
        % \multicolumn{2}{|c|}{Dataset Summary} \\
        \hline
        \multirow{2}{*}{\textbf{Model}} & \multirow{2}{*}{\textbf{Patch}} & \multicolumn{3}{|c|}{\textbf{im2latex}} & \multicolumn{3}{|c|}{\specialcell{\textbf{\ours} \\ \textbf{(Synthetic)}}} & \multicolumn{3}{|c|}{\specialcell{\textbf{\ours} \\ \textbf{(Real-World)}}} \\
        \cline{3-11}
         & \textbf{Size} & BLEU & Edit & EM & BLEU & Edit & EM & BLEU & Edit & EM  \\
        
        \hline 
        MI2LS-MLE & - & 89.08 & 91.09 & 79.39 & - & - & - & - & - & - \\
        \hline
        MI2LS-RL & - & \textbf{90.28} & \textbf{92.28} & \textbf{82.33} & - & - & - & - & - & - \\
        
        \hline
        \hline 

        % \hline 
        OCR-ViT & 10 & \textbf{84.53} & \textbf{87.45} & \textbf{36.92} & \textbf{99.53} & \textbf{99.64} & \textbf{98.31} & \textbf{81.24} & \textbf{86.09} & \textbf{51.05} \\
        \hline
        OCR-ViT & 16 & 72.11 & 77.11 & 19.48 & 98.76 & 99.20 & 94.85 & 72.52 & 77.96 & 41.96 \\
        \hline 
        \specialcell{OCR-ViT\\(w/o bolding,\\pixelation, padding)} & 16 & 65.59 & 71.57 & 20.91 & 98.85 & 99.25 & 99.13 & 55.27 & 58.73 & 12.24 \\
        \hline 
        \specialcell{OCR-ViT (im2latex \textbf{only})} & 16 & 31.11 & 39.06 & 0.83 & 0.42 & 2.45 & 0.01 & 0.68 & 3.38 & 0.00 \\
        \hline 
        \specialcell{OCR-ViT (\ours \textbf{only})} & 16 & 1.60 & 13.42 & 0.00 & 98.89 & 99.28 & 95.09 & 67.79 & 74.60 & 37.06 \\

        \hline
        \hline
        
        \specialcell{Pix2Tex (im2latex \textbf{only})} & 8 & \textbf{90.24} & \textbf{91.78} & \textbf{39.24} & 0.97 & 4.03 & 0.03 & 2.90 & 8.84 & 0.35 \\
        \hline
        \specialcell{Pix2Tex\\(\textbf{im2latex + 10\% \ours}, \\ w/o bolding, pixelation,\\padding)} & 8 & 87.95 & 89.44 & 33.24 & \textbf{99.19} & \textbf{99.38} & \textbf{94.93} & \textbf{68.85} & \textbf{74.74} & \textbf{31.82} \\
        \hline
    \end{tabular}
    \caption{Performance of various model architectures trained using different patch sizes on im2latex, \ours, and \ours Real-World test sets. 
    % The test sets of im2latex and printed English are utilized here. 
    For all three of our performance metrics, a higher value indicates a better model.
    Unless stated otherwise, models were trained on \textit{both} im2latex and \ours.
    }
    \label{tab:patch_size_results}
\end{table*}
\setlength\tabcolsep{6pt}

\subsection{Effect of Patch Size}
The performance of training our OCR-ViT model with two different patch sizes, 10 and 16 are included in Table \ref{tab:patch_size_results}. We see that a lower patch size yields stronger performance in all cases, but these performance improvements are most pronounced on im2latex and the \ours real-world test set. Considering the discussion in Section~\ref{sec:impact_of_patch_size}, this makes sense. The records in im2latex and \ours real-world are more \textit{complex} than vanilla printed English, which comprises the majority of the synthetic \ours test set. As a consequence, each \textit{patch} is burdened with expressing relatively \textit{more} information in these settings, and lowering the patch size lowers thus burden. This advantage of lower patch size we find on OCR-ViT facilitates us to pursue a patch size of 8 on Pix2Tex. % This is the smallest value we have tried on Pix2Tex.

\subsection{Multi-Domain Training}
In this section, we present results from our experiments exploring the impact of multi-domain training for domain-specific inference. 
That is, we explore the effect of training on the union of im2latex-100k and \ours versus only on im2latex-100k. 
% The test set of im2latex-100k and synthetic \ours are utilized for model evaluation. The \ours real-world test set is also adopted. 
Performance is evaluated on three test sets: im2latex-100k, \ours synthetic test set, and \ours real-world test set.

For the OCR-ViT models, performance on \ours synthetic test set is relatively unchanged as long as \ours is included in the training corpus - in isolation \textit{or} with im2latex-100k. Performance on im2latex-100k and \ours real-world test set is greatly improved when a model is trained on both corpora than either one in isolation. We see that multi-domain training yields an average improvement of 825.39\% on im2latex (primarily in EM) and 8.23\% on \ours real-world.

For Pix2Tex, comparing the last two rows of Table~\ref{tab:patch_size_results}, we see that adding \ours into training yields substantial performance boost on \ours synthetic and real-world test set. Scores on im2latex-100k are slightly lower than training on im2latex-100k alone, but it is clear that multi-domain training yields the best overall performance on Pix2Tex model. This intuitively makes sense, as our target domain is a hybrid of printed English and scientific text. The advantage of multi-domain training reconfirms the value of proposing \ours, since constructing a hybrid dataset like \ours is not straightforward using existing resources due to reasons such as inconsistency of datasets across various domains. Note that we use 10\% of \ours without bolding, pixelation, and padding to reduce training cost, as it would take roughly three months to train on the entirety of \ours using the script provided by Pix2Tex.

\subsection{Significance of Real-World Test Set}
In Table \ref{tab:patch_size_results},
% , \ref{dataset_tab}, and \ref{pretrained} in aggregate, 
it is clear to see that all the models yield significantly worse performance when they are tested on \ours real-world test set compared to when they are tested on other data. Since this real-world test set 
% does not contain pseudo-chemical equations and is meant to mimic the real-world scenarios,
is comprised of artifacts from \textit{real} scientific documents,
we conclude that it helps to show the actual capability of each OCR model on real-world chemistry scholarly papers. This novel test set clearly reveals the weakness of each selected model, so its value is demonstrated.

\subsection{Impact of Record Transformations}
In comparing rows 4 and 5 of Table \ref{tab:patch_size_results}, we see that our proposed bolding, pixelation, and padding transformations yield performance improvements on the im2latex and \ours real-world test sets, with most pronounced improvements on the \ours real-world test set. The transformations yield an \textit{average} improvement of 3.61\% and 102.25\% on im2latex test set and \ours real-world test set, respectively, in BLEU, edit distance, and exact match. Performance on \ours synthetic test set is largely unchanged with and without the transformations.

Intuitively, these results make sense. The impact of the transformations is most pronounced on the type of records they were designed to mimic, but they increase the model's performance in other applications also. These transformations lead to a slight performance decrease on the \ours synthetic test set, but an overall stronger model. 

\section{Conclusion}

In this paper we introduce the \ours dataset for OCR, containing more than 1M (rendered text, \LaTeX{} ground-truth) pairs of printed English and chemical equation text. The dataset contains three subsections: printed English, pseudo chemical equations, and images of text extracted from real-world scientific documents. This dataset helps bridge the gap between OCR models and datasets designed for either vanilla printed English or scientific text (\textit{e.g.}, math and physics formulae), but not both. 

Additionally, we survey a variety of architectures when applied to the \ours dataset. 
% We ultimately find that Pix2Tex, a transformer-based encoder-decoder model, augmented with a CNN encoder, yields the best performance. Pix2Tex beats the previous state-of-the-art on the im2latex test set, MI2LS, in BLEU score and edit distance, falling short in the exact match metric. 
We find that a traditional ViT with small patch size ($10*10$), trained in a multi-domain setting using our proposed pixelation, bolding, and padding transformations yields the best \textit{overall} performance. However, a Pix2Tex model (\textit{i.e.} ViT + CNN encoder) yields competitive performance when trained on only 10\% of \ours \textit{without} our proposed transformations
% - surpassing a vanilla OCR-ViT on im2latex - 
suggesting a promising path forward for future OCR models. 

% Finally, in comparing the effect of different patch sizes in the ViT, we see that smaller patch sizes lead to improved performance in OCR applications, contradicting previous results in general CV. As a consequence, the transformer must process a longer effective sequence of patches, incurring higher computational overhead. However, similar levels of performance can be obtained at lower computational cost using a \textit{larger} patch size but first passing the source image through a ResNet encoder before patching. For the impact of multi-domain training, although it may not help domain-specific inference, it is a necessary step for building OCR models on scientific documents.

\section{Acknowledgements}
This material is based upon work supported by the U.S. Department of Energy’s Office of Energy Efficiency and Renewable Energy (EERE) under the Advanced Manufacturing Office Award Number DE-EE0007897 awarded to the REMADE Institute, a division of Sustainable Manufacturing Innovation Alliance Corp.

This report was prepared as an account of work sponsored by an agency of the United States Government. Neither the United States Government nor any agency thereof, nor any of their employees, makes any warranty, express or implied, or assumes any legal liability or responsibility for the accuracy, completeness, or usefulness of any information, apparatus, product, or process disclosed, or represents that its use would not infringe privately owned rights. Reference herein to any specific commercial product, process, or service by trade name, trademark, manufacturer, or otherwise does not necessarily constitute or imply its endorsement, recommendation, or favoring by the United States Government or any agency thereof. The views and opinions of authors expressed herein do not necessarily state or reflect those of the United States Government or any agency thereof.

\nocite{*}
\section{Bibliographical References}\label{sec:reference}
\vspace{-22.5pt}

\bibliographystyle{lrec-coling2024-natbib}
\bibliography{lrec-coling2024-example}

% \section{Language Resource References}
% \label{lr:ref}
% \bibliographystylelanguageresource{lrec-coling2024-natbib}
% \bibliographylanguageresource{languageresource}

\end{document}